# A reinforcement learning based construction material supply strategy using robotic crane and computer vision for building reconstruction after an earthquake

Yifei Xiao[1], T.Y. Yang[2], Xiao Pan[3], Fan Xie[4] and Zhongwei Chen[4]

[1] *Ph.D. student, Department of Civil Engineering, University of British Columbia, yifeix5@student.ubc.ca*
[2] *Professor, Department of Civil Engineering, University of British Columbia, yang@civil.ubc.ca*
[3] *Research fellow, Department of Civil Engineering, University of British Columbia, xiao.pan@civil.ubc.ca*
[4] *Master student, Department of Civil Engineering, University of British Columbia*

## 1. ABSTRACT

After an earthquake, it is particularly important to provide the necessary resources on site because a large number of infrastructures need to be repaired or newly constructed. Due to the complex construction environment after the disaster, there are potential safety hazards for human labors working in this environment. With the advancement of robotic technology and artificial intelligent (AI) algorithms, smart robotic technology is the potential solution to provide construction resources after an earthquake. In this paper, the robotic crane with advanced AI algorithms is proposed to provide resources for infrastructure reconstruction after an earthquake. The proximal policy optimization (PPO), a reinforcement learning (RL) algorithm, is implemented for 3D lift path planning when transporting the construction materials. The state and reward function are designed in detail for RL model training. Two models are trained through a loading task in different environments by using PPO algorithm, one considering the influence of obstacles and the other not considering obstacles. Then, the two trained models are compared and evaluated through an unloading task and a loading task in simulation environments. For each task, two different cases are considered. One is that there is no obstacle between the initial position where the construction material is lifted and the target position, and the other is that there are obstacles between the initial position and the target position. The results show that the model that considering the obstacles during training can generate proper actions for the robotic crane to execute so that the crane can automatically transport the construction materials to the desired location with swing suppression, short time consumption and collision avoidance.

Keywords: Robotic crane, earthquake engineering, reinforcement learning, 3D path planning, motion planning.

## 1. INTRODUCTION

Earthquake damage to infrastructure can result in casualties, economic losses and functional losses. After an earthquake, a large number of infrastructures need to be repaired or newly constructed. In traditional methods, human labors will operate heavy equipment, such as tower cranes or mobile cranes, on the construction site to provide the necessary construction materials. However, due to the complex construction environment after the disaster, there are potential safety hazards for human labor working in this environment. In this case, robotic technologies with artificial intelligent (AI) algorithms could be the potential solution to provide construction resources after an earthquake. In post disaster reconstruction, mobile cranes are indispensable for transporting, handling and cleaning materials. They can be not only used for erecting structural members but are also used for lifting and transporting precast structural components, non-structural components and other materials in construction sites. Since a great number of construction activities rely on the efficiencies of resource transportation by cranes, the proper planning and operations of cranes are essential to achieve high productivity during construction. The lift path planning of cranes is one of the important tasks when operating a crane. Currently, mobile cranes are normally controlled by crane operators through a control system attached to the crane itself or a wireless remote controller. Therefore, the lift path is entirely judged and planned by crane operators. However, finding an optimal lift path with short time consumption is difficult for crane operators, and controlling the swing of the hoist rope on the crane is also a tricky problem to tackle when transporting construction materials. With the advancement of robotic technologies and machine learning (ML), new opportunities have been brought to the construction industry. The robotic crane with AI algorithms can be implemented to replace human beings for lifting, transporting and construction tasks after an earthquake.

With the advancement of robotic and control technologies in civil, structural and construction engineering [1-4, 17]; Xiao et al., 2023), research on the application of robots in construction has been investigated recently. Unmanned aerial vehicle (UAV) and unmanned ground vehicle (UGV) have been implemented for simple construction tasks. However, UAV can only perform the tasks that require lifting low-weight structural components due to its limited payload and flight time [5-7], and UGVs were mainly implemented to conduct construction of planar structures [8-9]. Robot arm, which is more powerful than UAV and UGV in construction, has also attracted researchers' interests these years. Koerner et al. [10] proposed an efficient workflow considering design, fabrication, and assembly for Digital and robotic automation. The KUKA robotic arm was used to assemble a discrete timber structure to achieve digital fabrication and robotic assembly. Liang et al. [11] implemented learning from demonstration (LfD) to teach robotic arm how to install ceiling tile using a set of human demonstration videos. The robotic arm has been proved to have high degrees of freedom and can be used to complete the assembly of small-scale structures. However, they cannot be implemented for large-scale and heavy-weight structure construction [12]. Compared to robotic arm, robotic crane is more suitable for lifting, transporting and installing large-scale and heavy-weight structural components. Research on robotic cranes has also been attempted for autonomous construction. For example, Zi et al. [13] proposed a solution to address the cooperative problems, including localization, obstacle avoidance and automatic leveling control, for four mobile cranes. Dutta et al. [14] employed intelligent decision-making and planning algorithms to establish a lift planning system for tower cranes with the support by building information modeling (BIM). Kayhani et al. [15] developed a heavy-lift path planning method to find the shortest path for the crane planar motion.

Machine learning (ML) has been applied in various applications in civil, structural and construction engineering due to the advancement of AI technology. In the field of civil and structural engineering, Deep Learning (DL, a type of ML algorithms) has achieved a great success in a wide range of computer vision tasks [16-17], such as image classification [18-19], object detection [20-21], segmentation [22], motion tracking [23-24], and 3D vision-based applications [25-26, 35]. In construction engineering, DL has been applied to recognize actions of onsite workers or equipment on the construction site [27-29].

On the other hand, reinforcement learning (RL), one of the other powerful ML algorithms, has been widely investigated in construction recently. Sartoretti et al. [30] employed distributed asynchronous advantage actor-critic (A3C) algorithm to learn a policy that allowed multiple agents to work together to achieve a same goal. The simulated results showed that multiple agents were trained to be able to work in a shared environment to complete a collective construction task. Lee and Kim [31] investigated the feasibility of deep Q-network (DQN) for autonomous hoist control. The results proved that the implementation of DQN can increase the lifting efficiency and reduce unnecessary trips when multiple hoists are operating simultaneously. Jeong and Jo [32] used a convolutional neural network based deep deterministic policy gradient (DDPG) approach to train the RL agent to be able to design a reinforced concrete beam in a cost-effective way. These studies have proved that the implementation of RL can effectively improve the efficiency in designing structures, transporting materials and conducting construction tasks on the construction site. However, more research could be investigated for the implementation of RL in mobile crane motion planning.

In this paper, a robotic crane is used to provide resources for infrastructure reconstruction after an earthquake. The proximal policy optimization (PPO) [33], a RL algorithm, is implemented to train a model that can conduct 3D lift path planning for a robotic crane. The state and reward function are designed in detail for RL model training. Two models are trained through a loading task in different environments: 1) model #1 considers the influence of obstacles between the initial position where the construction material is lifted and the target position; 2) model #2 only considers the robotic crane itself and ground as obstacles (no obstacles between the initial position and the target position). After training, the performance of the two trained models is compared through a loading task and an unloading task in a simulation environment. The loading task is to transport the construction materials from the ground to the designated location on crane, and the unloading task is to unload the construction materials from the crane and transport it to the designated location on the ground. For each task, two cases, whether there are obstacles between the construction material initial position and the target position, are investigated and compared. The simulated results proved that the model #1 is successfully trained through PPO algorithms that allows robotic crane to automatically transport the construction materials to the desired location with swing suppression, short time consumption and collision avoidance.

## 2. METHODOLOGY

### 2.1 PPO and neural network parameter update

In this section, an on-policy reinforcement algorithm, PPO, is employed to train the agent to learn a policy that allows a robotic crane to automatically plan a lift path and transport construction materials with swing suppression, short time consumption and collision avoidance. Since the lift path planning of the robotic crane is a continuous action space planning problem, the PPO- continuous algorithm is adopted, where the Gaussian distribution is used for the output of crane action. There are two neural networks in the PPO algorithm, which are actor network and critic network. The actor network is used to generate actions based on the states obtained from environment while the critic network is used to evaluate the values of the states. The parameters of the actor network are updated based on the Equation (1):

$$\theta_{k+1} = \arg\max E\left[\frac{\pi_\theta(a|s)}{\pi_{\theta_{old}}(a|s)} A^{\pi_{\theta_{k+1}}}(s,a), clip\left(\frac{\pi_\theta(a|s)}{\pi_{\theta_{old}}(a|s)}, 1-\varepsilon, 1+\varepsilon\right) A^{\pi_{\theta_{k+1}}}(s,a)\right] \quad (1)$$

where $\pi$ is the policy network (actor network) and $\theta$ is the parameters of actor network. $a$ is the action taken by the agent and $s$ is the state obtained from the environment. $A(s,a)$ is the advantage function. $\varepsilon$ is a small hyperparameter. The parameters ($\varphi$) of the critic network are updated based on the Equation (2):

$$\varphi_{k+1} = \arg\min \frac{1}{|D_k|T} \sum_{\tau \in D_k} \sum_{t=0}^{T} \left(V_\varphi(s) - r\right)^2 \quad (2)$$

where $D_k$ is a set of trajectories. $T$ is the total steps. $V_\varphi(s)$ is the value function and $r$ is the reward function.

### 2.2 Design of state

The states observed from the environment are the inputs of the actor network and critic network. The quality of the states will affect the result of the agent training and the convergence of PPO algorithm. In order to train the agent to complete the loading and unloading tasks, the states should be designed based on the requirements of the task. Table 1 summarizes the states used in this study when training the agent through PPO algorithm.

*Table 1 Summary of states for PPO training*

| State | Description | Value |
|---|---|---|
| Position of the material | The real-time position of the material lifted at each time step. | 3D coordinates of the material lifted |
| Target position | The position of the target destination. The position is only updated at the initial stage of each episode. | 3D coordinates of the target position |
| Distance | The distance between the material lifted and the target position. The distance is updated after the robotic crane executes a new action output by actor network | Distance in 3D space |
| Collision warning | A lidar sensor with 5m distance range is installed on the hook to detect the obstacle | 0, 1 |
| Rope angle | The angle of hoist rope swing (should be smaller than the maximum allowable swing angle $\theta_{thr}$) | 0° - $\theta_{thr}$ |
| Steps | Current step consumption in each episode (should be smaller than the maximum allowable steps $N_{step}$ of each episode) | 0 - $N_{step}$ |

**2.3 Design of reward function**

The target of learning a policy using RL algorithm is normally to train a RL model that can maximize the future total reward. The design of the reward function can determine whether the policy can output a proper action according to the input states. A properly designed reward function can guide the agent to learn a policy that promotes the agent to complete the task successfully, while an incorrect reward function will result in the failure of the task. In this paper, several continuous rewards and discrete rewards are designed to guide the training of RL model. The designed reward functions are listed in Table 2.

*Table 2 Design of reward function*

| Reward No. | Reward Type | Description | Value |
|---|---|---|---|
| 1 | continuous rewards | positive reward for getting closer to the target position | $p_1 \times 2^{-|L|}, (p_1 > 0)$ |
| 2 | continuous rewards | negative reward for swing of the hoist rope | $p_2 \times \frac{\theta_{rope}}{\theta_{thr}}, (p_2 < 0)$ |
| 3 | continuous rewards | negative reward for additional time consumption | $p_3, (p_3 < 0)$ |
| 4 | discrete rewards | negative reward for collision | $p_4, (p_4 < 0)$ |
| 5 | discrete rewards | positive reward for reaching target position | $p_5, (p_5 > 0)$ |

Note: $p_1$ to $p_5$ are the parameters defining the magnitute of each reward. $L$ is the distance between the material position and target position. $\theta_{rope}$ is the swing angle and $\theta_{thr}$ is the maximum allowable swing angle

**2.4 Modelling strategies**

The PyBullet [34] is used as the simulation software for modelling and RL model training. All models in the simulation environment are created using universal robot description file (URDF) format codes. Figure 1 shows the URDF model of robotic crane. To make the trained model more practical, several modelling strategies are used in this study: 1) all models including robotic crane, construction material and obstacles are built as 1:1 scale models; 2) the robotic crane is modelled according to the real specifications of the crane COPMA 510_ENG; 3) the hoist rope is created by a number of rigid links and continuous type joints. In addition, several reasonable assumptions are also made during modelling: 1) the actions output by the action network are increments of 3D coordinates of the end of telescopic boom; 2) the joint between the turntable and telescopic boom, whose rotation is normally driven by an actuator, is modelled as a revolute joint; 3) all the joint movements of the robotic crane are continuous and simultaneous; 4) a virtual lidar sensor is installed on the hook to detect the obstacle within 5m.

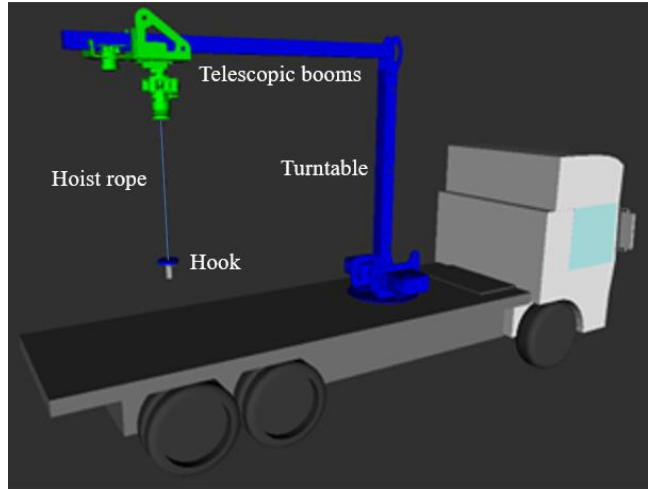

*Figure 1 URDF model of robotic crane*

## 3. MODEL TRAINING AND VALIDATION

In this section, the training of RL model for lift path planning is conducted through a loading task. Two RL models are trained in situations with and without obstacles respectively. Then, the two trained models are compared and evaluated through an unloading task and a loading task to check whether the proposed strategy can provide construction materials for a damaged structure after an earthquake. For each task, two different cases are considered. One is that there is no obstacle between the initial position where the material (a timber column) is lifted and the target position, and the other is that there are obstacles between the initial position and the target position.

### 3.1 Model training in a loading task

During training, a 5m-long cylinder timber column needs to be lifted and transported from the initial position at the ground to the target position on robotic mobile crane. Figure 2(a) and Figure 2(b) show the loading environment with and without the obstacles, respectively. In order to make sure that the timber column can be transported to any designated location rather than a fixed location, the target position is randomly assigned within the white rectangle at the beginning of each episode of training. The red rectangle represents the target position where the center of the cylinder timber column needs to reach. The initial position on the ground of the timber column is fixed during training. The 3D lidar scanner is usually used to scan the working environment of the construction site. Therefore, in the situation where the obstacle is considered when training the RL model, two lidar scanners supported by a desk are considered as the obstacle between the initial position and the target position.

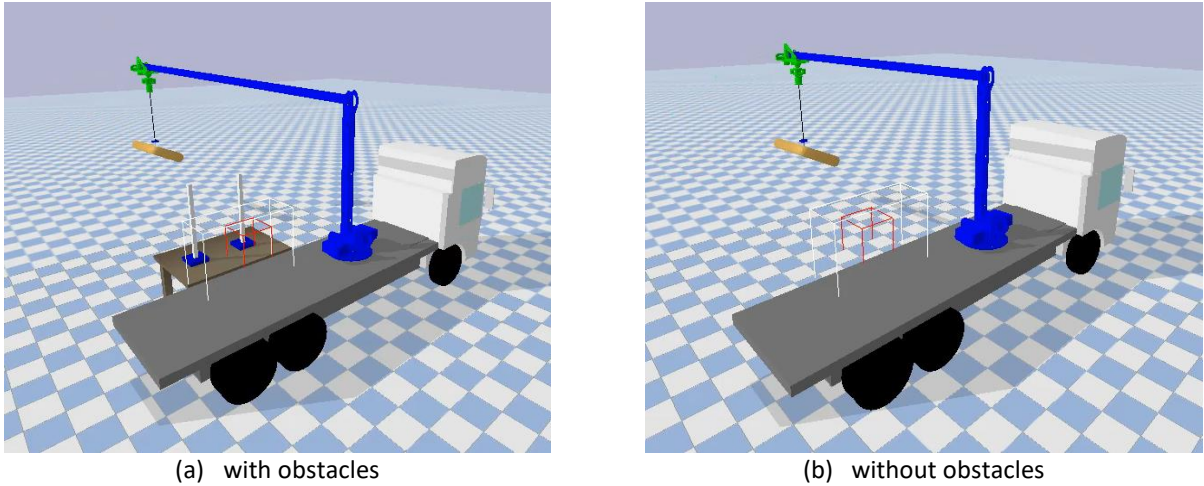

(a) with obstacles  (b) without obstacles

*Figure 2 Loading environment for PPO training*

For both cases with and without obstacles, a total of 1 million steps are set for policy training and the learning rate is set to 3×10^5. Figure 3 plots the cumulative reward curve for loading task throughout training. The results show that the cumulative rewards for two cases converge to the maximum level as the number of steps increases, which indicates the success of training for both two cases. It can also be observed that the training result converges faster in Case 2. This is because the training environment is more complicated in case 1 due to the obstacles between the initial position and target position. It is also noticeable that the cumulative rewards in case 1 have a slight decrease and then quickly re-converge to the maximum value after first convergence to the maximum reward level. This phenomenon can sometimes be solved by reducing the learning rate during training.

Figure 4 and Figure 5 show the loading procedures after training. The black line drawn in Figure 4(a) and Figure 5(a) represents the lift path of the object being transported to the target position. By comparing the lift path drawn in Figure 4(a) and Figure 5(a), it is evident that the crane automatically plans a lift path that first transports the timber column to a point above the two lidar scanners and then transports to the target positions when obstacles are considered. On the other hand, the timber column is directly transported to the target position when there is no obstacle.

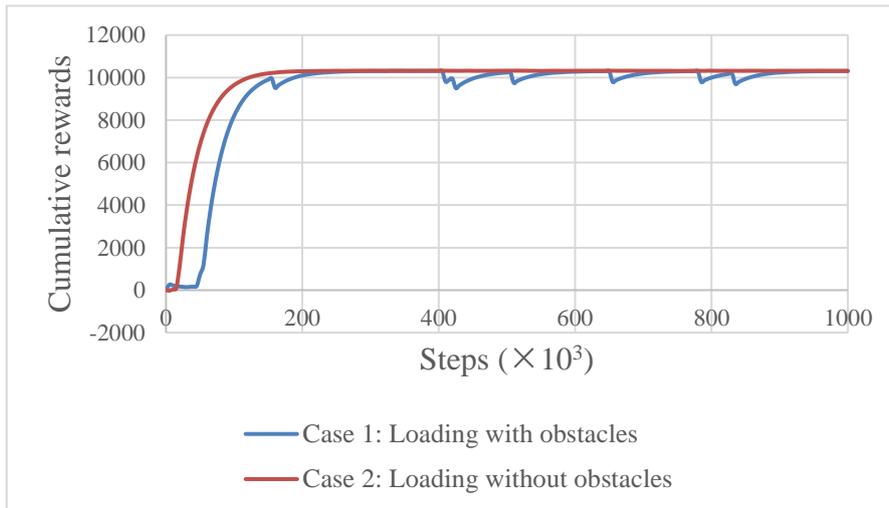

*Figure 3 Learning curve of loading task*

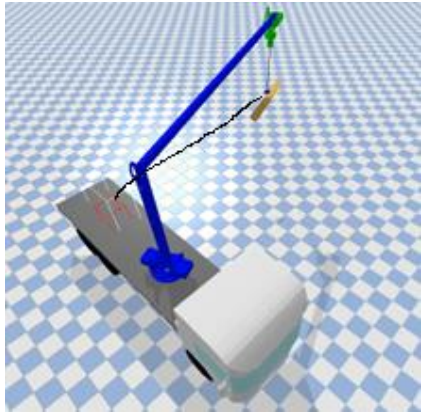
(a)

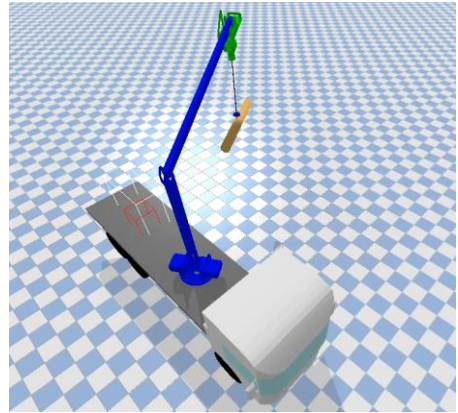
(b)

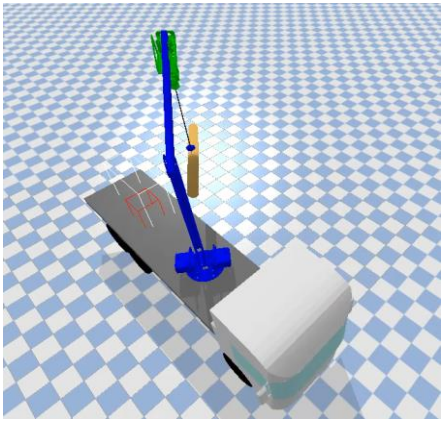
(c)

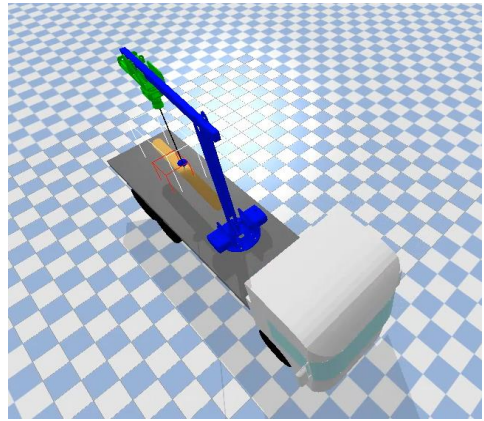
(d)

*Figure 4 Loading procedures without obstacles*

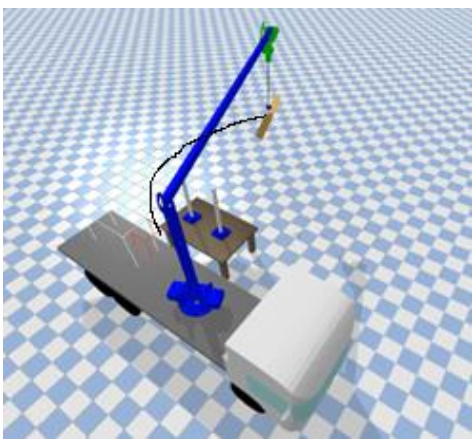
(a)

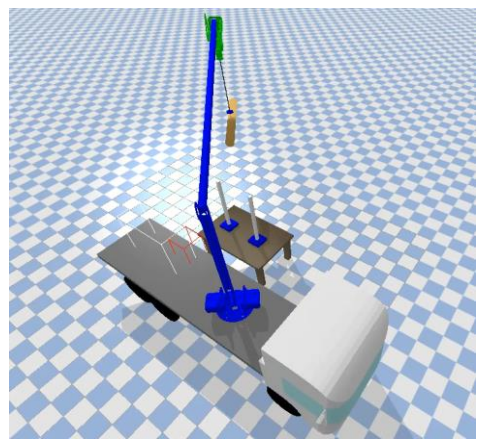
(b)

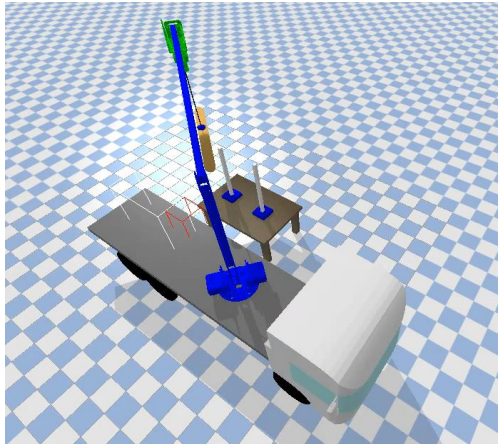 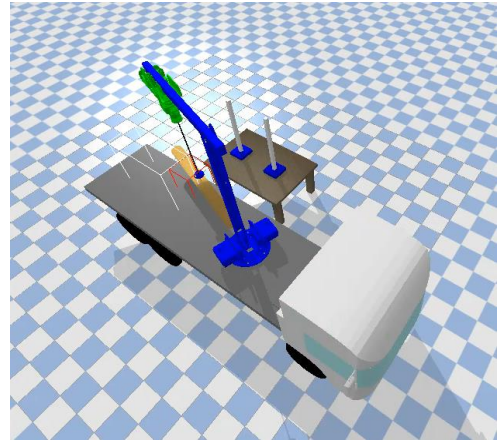

(c)                                                             (d)

*Figure 5 Loading procedures with obstacles*

### 3.2 Model evaluation

In this section, the two well-trained RL models are compared and evaluated through an unloading task and a loading task. The model trained considering lidar sensor as the obstacle is model #1 and the model trained without considering the obstacle is model #2. For both unloading task and loading task, the obstacle and obstacle-free situation are all considered. For the situation where the obstacle is considered, a wooden fence is added between the 5m-long cylinder timber column and the target position. In addition, a damaged wooden structure which requires reconstruction after an earthquake and a pre-scanned ground are imported into the PyBullet as the dynamic environment. During evaluation, a total number of 100 different scenarios with randomly selected target position are set for each task. The evaluation results of model #1 and model #2 are summarized in Table 3. Figure 6 and Figure 7 show the successfully examples of executing lift path planning task using model #1 and model #2, respectively. As can be seen from Table 3, model # 2 has similar performance compared to model #1 when the wooden fence is not added as an obstacle between the 5m-long cylinder timber column and the target position. However, model #1 has much better performance than model # 2 when the wood fence is considered as an obstacle. This is because model #1 has a more complex training environment where more obstacles are taken into consideration during the training process. Therefore, a RL model with superior performance needs to be obtained, more factors need to be considered during the training process. However, extremely complex environments can also increase the difficulty of training, and sometimes may even result in a failure of training. Hence, the training environment needs to be reasonably and carefully defined according to the task demands.

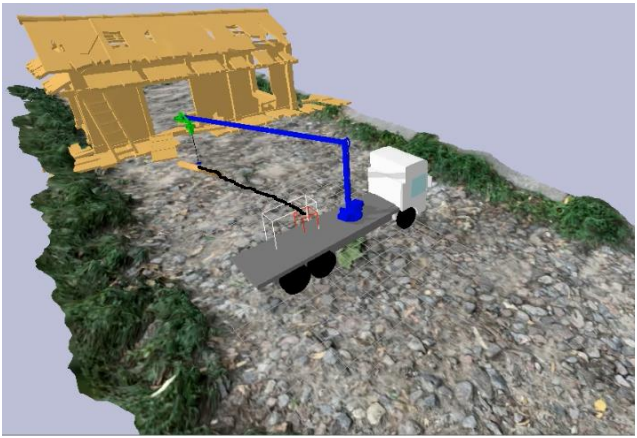 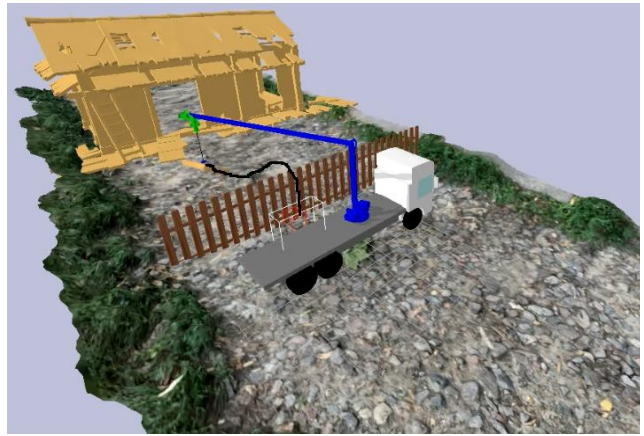

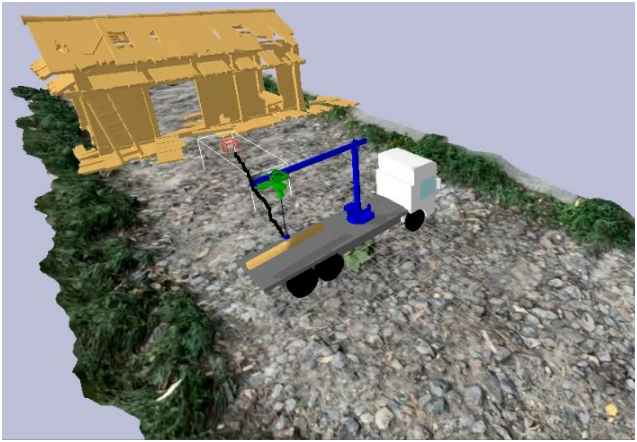
(a) loading task without obstacle

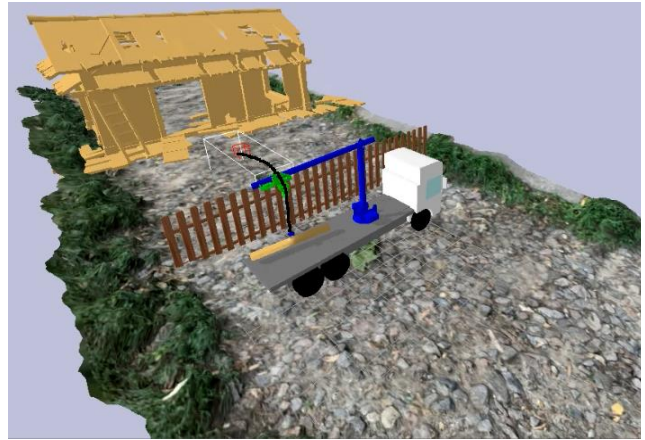
(b) loading task with obstacle

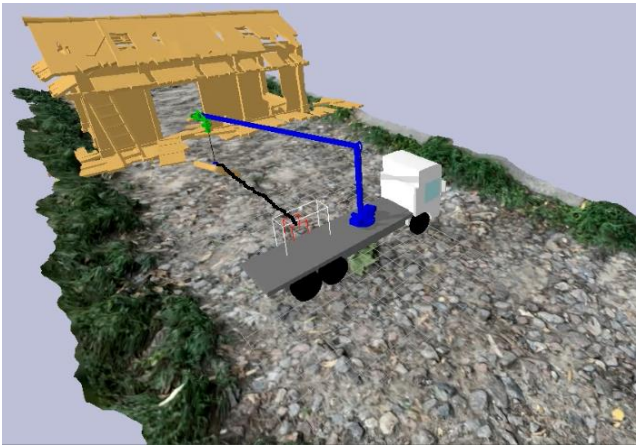
(c) unloading task without obstacle

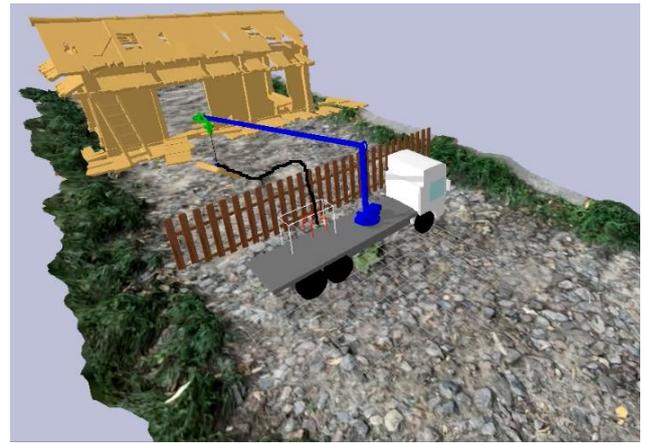
(d) unloading task with obstacle

*Figure 6 Successful examples of executing lift path planning task using model #1*

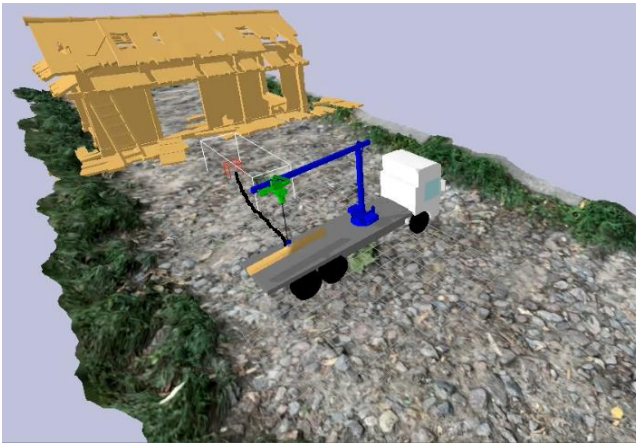
(a) loading task without obstacle

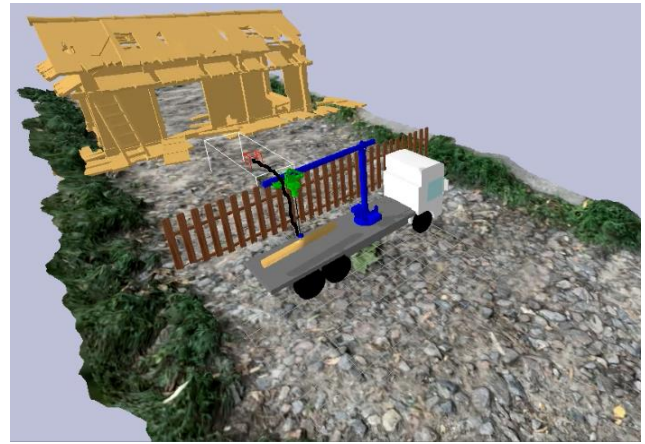
(b) loading task with obstacle

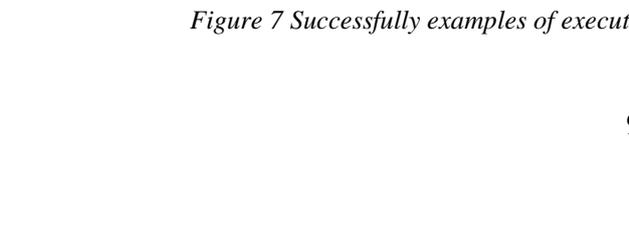
(c) unloading task without obstacle

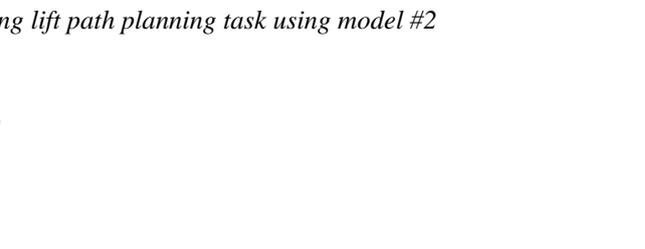
(d) unloading task with obstacle

*Figure 7 Successfully examples of executing lift path planning task using model #2*

*Table 3 Evaluation results of model #1 and model #2*

| Model # | Loading task without obstacle (%) | Loading task with obstacle (%) | Unloading task without obstacle (%) | Unloading task with obstacle (%) |
|---|---|---|---|---|
| 1 | 100 | 99 | 100 | 98 |
| 2 | 100 | 54 | 98 | 47 |

## 4. CONCLUSION

In this paper, the robotic crane with PPO is proposed to replace human labors to provide construction resources after an earthquake. A reinforcement learning based lift path planning is developed for construction material transportation. Two models are trained through PPO algorithms to automatically transport the construction materials to the target position with swing suppression, short time consumption and collision avoidance. The feasibility of PPO in lift path planning is validated through a loading task and an unloading task in a simulation environment. In each task, two cases, whether there are obstacles between the construction materials and the target position, are investigated and compared. The results show that model #1 can conduct proper lift path planning regardless of the existence of obstacles. Although RL has shown its potentials for lift path planning through this study, several limitations can be observed, and future studies will be conducted accordingly:

- The simulation environment in this paper is simple, more complicated environment should be investigated in the future.

- The PPO algorithm for lift path planning of robotic crane is only validated in a simulation environment, its feasibility still needs to be investigated in the laboratory and finally in a real construction site.

- The performance of other reinforcement learning algorithms, such as DQN, DDPG and soft actor-critic (SAC), should be investigated and compared with PPO algorithm.

- The implementation of reinforcement learning on construction tasks, such as installation of structural components and excavation, should also be investigated.

## 5. Reference

[1] Garcia, E., Jimenez, M. A., De Santos, P. G., & Armada, M. (2007). The evolution of robotics research. IEEE Robotics & Automation Magazine, 14(1), 90-103.
[2] Kang, S., Yu, J., Zhang, J., & Jin, Q. (2020). Development of multibody marine robots: A review. Ieee Access, 8, 21178-21195.
[3] Xiao, Y., Pan, X., & Yang, T. T. (2022). Nonlinear backstepping hierarchical control of shake table using high‐gain observer. Earthquake Engineering & Structural Dynamics, 51(14), 3347-3366.
[4] Korkmaz, S. (2011). A review of active structural control: challenges for engineering informatics. Computers & Structures, 89(23-24), 2113-2132.
[5] Lindsey, Q., Mellinger, D., & Kumar, V. (2011). Construction of cubic structures with quadrotor teams. Proc. Robotics: Science & Systems VII, 7.
[6] Kondak, K., Krieger, K., Albu-Schaeffer, A., Schwarzbach, M., Laiacker, M., Maza, I., Rodriguez-Castano, A., & Ollero, A. (2013). Closed-loop behavior of an autonomous helicopter equipped with a robotic arm for aerial manipulation tasks. International Journal of Advanced Robotic Systems, 10(2), 145. https://doi.org/10.5772/53754.

[7] Augugliaro, F., Lupashin, S., Hamer, M., Male, C., Hehn, M., Mueller, M. W., ... & D'Andrea, R. (2014). The flight assembled architecture installation: Cooperative construction with flying machines. IEEE Control Systems Magazine, 34(4), 46-64.

[8] Allwright, M., Bhalla, N., El-faham, H., Antoun, A., Pinciroli, C., & Dorigo, M. (2014, September). SRoCS: Leveraging stigmergy on a multi-robot construction platform for unknown environments. In International conference on swarm intelligence (pp. 158-169). Springer, Cham.

[9] Cucu, L., Rubenstein, M., & Nagpal, R. (2015, May). Towards self-assembled structures with mobile climbing robots. In 2015 IEEE International Conference on Robotics and Automation (ICRA) (pp. 1955-1961). IEEE.

[10] Koerner-Al-Rawi, J., Park, K. E., Phillips, T. K., Pickoff, M., & Tortorici, N. (2020). Robotic timber assembly. Construction Robotics, 4(3), 175-185.

[11] Liang, C. J., Kamat, V. R., & Menassa, C. C. (2020). Teaching robots to perform quasi-repetitive construction tasks through human demonstration. Automation in Construction, 120, 103370.

[12] Apolinarska, A. A., Pacher, M., Li, H., Cote, N., Pastrana, R., Gramazio, F., & Kohler, M. (2021). Robotic assembly of timber joints using reinforcement learning. Automation in Construction, 125, 103569.

[13] Zi, B., Lin, J., & Qian, S. (2015). Localization, obstacle avoidance planning and control of a cooperative cable parallel robot for multiple mobile cranes. Robotics and Computer-Integrated Manufacturing, 34, 105-123.

[14] Dutta, S., Cai, Y., Huang, L., & Zheng, J. (2020). Automatic re-planning of lifting paths for robotized tower cranes in dynamic BIM environments. Automation in Construction, 110, 102998.

[15] Kayhani, N., Taghaddos, H., Mousaei, A., Behzadipour, S., & Hermann, U. (2021). Heavy mobile crane lift path planning in congested modular industrial plants using a robotics approach. Automation in Construction, 122, 103508.

[16] Pan, X., & Yang, T. Y. (2022). Image‐based monitoring of bolt loosening through deep‐learning‐based integrated detection and tracking. Computer‐Aided Civil and Infrastructure Engineering, 37(10), 1207-1222.

[17] Xiao, Y., Pan, X., Tavasoli, S., M. Azimi, Noroozinejad Farsangi E., Yang T.Y. (2023) "Autonomous Inspection and Construction of Civil Infrastructure Using Robots." Automation in Construction Toward Resilience: Robotics, Smart Materials & Intelligent Systems, edited by Ehsan Noroozinejad Farsangi, Mohammad Noori, Tony T.Y. Yang, Paulo B. Lourenço, Paolo Gardoni Izuru Takewaki, Eleni Chatzi, Shaofan Li.

[18] Li, D., Cong, A., & Guo, S. (2019). Sewer damage detection from imbalanced CCTV inspection data using deep convolutional neural networks with hierarchical classification. Automation in Construction, 101, 199-208.

[19] Pan, X., & Yang, T. Y. (2020). Postdisaster image‐based damage detection and repair cost estimation of reinforced concrete buildings using dual convolutional neural networks. Computer‐Aided Civil and Infrastructure Engineering, 35(5), 495-510.

[20] Liang, X. (2019). Image‐based post‐disaster inspection of reinforced concrete bridge systems using deep learning with Bayesian optimization. Computer‐Aided Civil and Infrastructure Engineering, 34(5), 415-430.

[21] Pan, X. (2022). Three-dimensional vision-based structural damage detection and loss estimation–towards more rapid and comprehensive assessment (Doctoral dissertation, University of British Columbia). DOI: 10.14288/1.0422384.

[22] Tavasoli, S., Pan, X., Yang, T. Y. (2023). Real-time autonomous indoor navigation and vision-based damage assessment of reinforced concrete structures using low-cost nano aerial vehicles. Journal of Building Engineering. 106193.

[23] Shao, Y., Li, L., Li, J., An, S., & Hao, H. (2022). Target-free 3D tiny structural vibration measurement based on deep learning and motion magnification. Journal of Sound and Vibration, 538, 117244.

[24] Pan, X., Yang, T. Y., Xiao, Y., Yao, H., & Adeli, H. (2023). Vision-based real-time structural vibration measurement through deep-learning-based detection and tracking methods. Engineering Structures, 281, 115676.

[25] Pan, X., & Yang, T. Y. (2023a). 3D vision‐based out‐of‐plane displacement quantification for steel plate structures using structure‐from‐motion, deep learning, and point‐cloud processing. Computer‐Aided Civil and Infrastructure Engineering, 38(5), 547-561.

[26] Pan, X., Yang, T. Y. (2023b). 3D vision-based bolt loosening quantification using photogrammetry, deep learning, and point-cloud processing. Journal of Building Engineering. 106326.

[27] Yu, Y., Li, H., Yang, X., Kong, L., Luo, X., & Wong, A. Y. (2019). An automatic and non-invasive physical fatigue assessment method for construction workers. Automation in construction, 103, 1-12.

[28] Zhou, C., Xu, H., Ding, L., Wei, L., & Zhou, Y. (2019). Dynamic prediction for attitude and position in shield tunneling: A deep learning method. Automation in Construction, 105, 102840.

[29] Lakshmanan, A. K., Mohan, R. E., Ramalingam, B., Le, A. V., Veerajagadeshwar, P., Tiwari, K., & Ilyas, M. (2020). Complete coverage path planning using reinforcement learning for tetromino based cleaning and maintenance robot. Automation in Construction, 112, 103078.

[30] Sartoretti, G., Wu, Y., Paivine, W., Kumar, T. S., Koenig, S., & Choset, H. (2019). Distributed reinforcement learning for multi-robot decentralized collective construction. In Distributed Autonomous Robotic Systems: The 14th International Symposium (pp. 35-49). Springer International Publishing.

[31] Lee, D., & Kim, M. (2021). Autonomous construction hoist system based on deep reinforcement learning in high-rise building construction. Automation in Construction, 128, 103737.

[32] Jeong, J. H., & Jo, H. (2021). Deep reinforcement learning for automated design of reinforced concrete structures. Computer-Aided Civil and Infrastructure Engineering, 36(12), 1508-1529.

[33] Schulman, J., Wolski, F., Dhariwal, P., Radford, A., & Klimov, O. (2017). Proximal policy optimization algorithms. arXiv preprint arXiv:1707.06347.

[34] Coumans, E., Bai, Y. P., & PyBullet, A. (2016). A Python Module for Physics Simulation for Games, Robotics and Machine Learning.

[35] Pan, X., Tavasoli, S., Yang, T. Y. (2023). Autonomous 3D vision based bolt loosening assessment using micro aerial vehicles. Computer-aided Civil and Infrastructure Engineering.